\documentclass{svproc}
\usepackage{url}
\usepackage{amsmath}
\usepackage{graphicx}
\usepackage{float}
\usepackage{mathrsfs}
\usepackage{subfigure}
\usepackage[dvipsnames]{xcolor}
\usepackage{pgfplots}

\usepackage{color}

\pgfplotsset{compat=1.5}

\begin{document}
\mainmatter              
\title{Center-of-Mass-based Robust Grasp Pose Adaptation Using RGBD Camera and Force/Torque Sensing}
\titlerunning{CoM-Based Robust Grasp Pose Adaptation}  
%
\author{Shang Liu\inst{1}\footnotemark[1] \and Xiaobao Wei\inst{1}\footnotemark[1] \and Lulu Wang\inst{2} \and Jing Zhang\inst{1} \and Boyu Li\inst{3}\footnotemark[2] \and Haosong Yue\inst{1}}

\authorrunning{Shang Liu et al.} 
%
\tocauthor{Shang Liu, Xiaobao Wei, Songhao Yue}
\institute{School of Automation Science and Electrical Engineering, Beihang University, Beijing 100191, China \\ 
\and School of Mechanical and Electronic  Engineering, Northwest A\&F University, Xianyang 712100, China \\
\and Beijing Institute of Aeronautical Materials, Beijing 100095, China}

\maketitle              

\renewcommand{\thefootnote}{\fnsymbol{footnote}} 
\footnotetext[1]{These authors contributed equally to this work}
\footnotetext[2]{Corresponding author to provide e-mail: lizii@sina.com}

\begin{abstract}
Object dropping may occur when the robotic arm grasps objects with uneven mass distribution due to additional moments generated by objects' gravity. To solve this problem, we present a novel work that does not require extra wrist and tactile sensors and large amounts of experiments for learning. First, we obtain the center-of-mass position of the rod object using the widely fixed joint torque sensors on the robot arm and  RGBD camera. Further, we give the strategy of grasping to improve grasp stability. Simulation experiments are performed in "Mujoco". Results demonstrate that our work is effective in enhancing grasping robustness.
\end{abstract}
\section{INTRODUCTION}

Grasping or lifting strategies for object manipulation have been studied in the past few years, especially for picking up objects of different masses or shapes. In the most recent cutting-edge research, the main proposal can allow robots to perform a grasping task for a large variety of daily objects \cite{GPD}. However, the paramount consideration in generating manipulation strategies is objects' geometry while ignoring the more common real-life situations where the mass is not uniformly distributed. A reasonable grasping strategy should consider the intrinsic properties of the object, such as mass distribution, especially the location of the object's center of mass(CoM). On the one hand, the closer the robot's grasping position is to the CoM, the more robust the grasping could be, so that the object is less likely to fall off during the process of robot manipulating. On the other hand, this strategies can make the torque generated by the joints on the robot arm as small as possible to avoid arm damage. Therefore, the extrinsic properties of the object, such as geometry, and intrinsic properties, such as CoM need to be taken into account when we generate grasping pose.

Recently, some studies have been conducted on adjusting the above grasping strategies through interaction between robots and the environment. However, these works require many trials to collect data for training in different working platforms that cannot be widely used.

To improve the robustness of grasping  during manipulation, we present a novel method that does not require additional wrist sensors or large amounts of experimental data. The CoM position of the rod object can be obtained by using the widely fixed joint torque sensors on the robot arm and the RGBD camera. Further, we give grasping strategies based on the CoM position. We perform simulation experiments in "Mujoco," and it demonstrates that our work effectively improves the success rate of grasping.

\section{RELATED WORK}

A large number of existing works generate grasp positions based on RGBD images of objects for robotic arms to perform grasp actions. These works can be mainly divided into planar grasp (3D) and 6D grasp depending on the task categories. Planar grasp means that the object is constrained to a working surface, and the gripper is only capable of grasping from only one direction, while 6D grasp means that the gripper can grasp at any angle and position in 3D space. In this paper, we mainly focus on 2D grasp. For 2D grasp, A rectangular representation is proposed, which can generate grasp positions for a wide range of objects \cite{Efficient-grasping}, and the introduction of deep learning methods further improves the success rate of grasping \cite{DL-Detecting-Robotic-Grasps}. However, for objects with inhomogeneous density distribution, grasping based only on the geometric shape will likely lead to grasping failures due to the offset of grasping positions and the object's CoM.

By interacting with the object through the robot arm, some work can also obtain other characteristics of the object besides its geometric shape, such as mass, center of gravity position, and surface friction of the object \cite{Inertial-Properties}, which can help us to handle more uncertainty in the grasping task. For example,  the force and moment information obtained from the wrist of the machine can be used to calculate the weight of the object and the position of the CoM \cite{4048449}. The time-series data collected by the touch sensor when the robot grasps the object can also be used to analyze the friction of the object surface \cite{slip-detection-strategies} or grasping stability based on 2-D time data sequence processing method\cite{Slip-Detection}.

To obtain better grasp stability, some works make the next grasp as close to the object CoM as possible based on the sensor data obtained from the first grasp. In order to minimize the force on the wrist of the robot arm when grasping a rod-like heavy object, a method is proposed to adjust the grasping position based on the feedback of the wrist moment of the robot arm \cite{center-of-mass}. However, this work does not consider the possibility of object sliding relative to the jaws during grasping. A deep learning method based on the wrist force and touch sensor data makes the arms grasp close to the CoM of the object at the second grasp by one trial grasp \cite{Center-of-Mass-based}. However, this work requires a large number of trials to collect data for training in different working platforms and cannot be widely used.

Our proposed algorithm does not require additional wrist moment/tactile sensors or additional acquisition of large amounts of experimental data. We can obtain the CoM position of the rod object using the widely fixed joint torque sensors on the robot arm and the RGBD camera on the side of the robot jaw. The CoM position estimation for arbitrary objects is a future research direction.

\section{PROBLEM STATEMENT}
We aim to achieve a stable grasp pose for rod-shaped objects with uneven mass distribution using a sensor-equipped n-DoF robotic arm with a parallel-jaw gripper. If the gripper grasps the geometric center of the object, the offset between the object CoM and the grasping position will generate additional gravitational moment that is likely to cause object slipping. Therefore, we want to estimate the position of the object's CoM based on the first trial grasp so that the second grasp can grasp as close to the object's CoM as possible.Figure 1 illustrates our pipline.
\begin{figure}[h]
\includegraphics[width=0.6\textwidth]{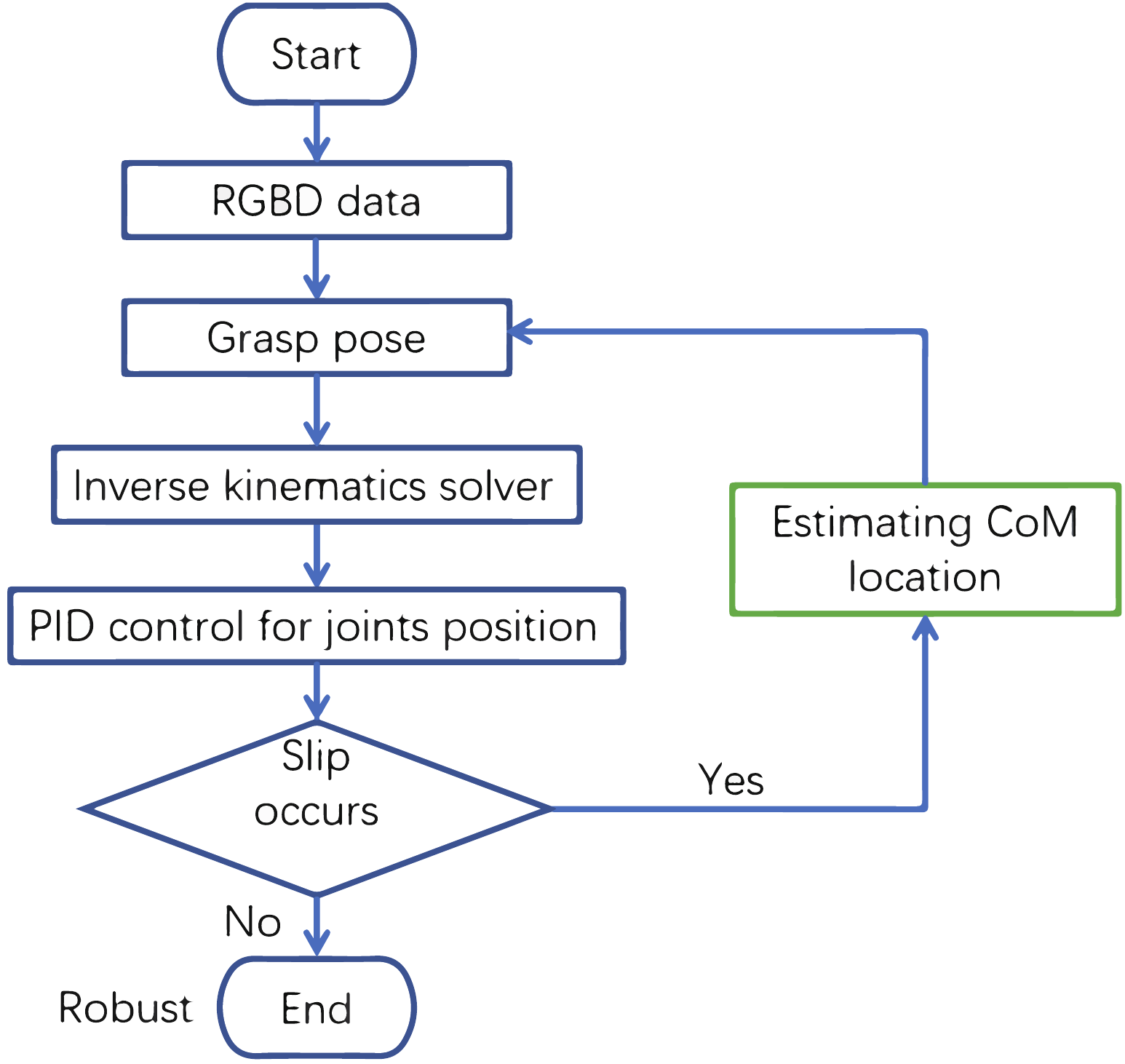}
\centering
\caption{The pipeline of our proposed method }
\label{fig:the pipeline of our proposed method}
\end{figure}

\section{METHODS}

\subsection{Object posture estimation}
In this section, we provide methods for geometry-based grasping pose generation and object posture identification which is very important for estimating the CoM location in the object frame.

We use two RGBD cameras: the Top-down camera and Side camera. The Top-down camera is placed directly above the operating table and its optical axis direction is parallel to the gravity direction. The Side camera is fixed on the end of the robotic arm (eelink) so that it can move with eelink and its optical axis is horizontal. The Top-down camera is used to identify object's $(x,y)$ position and rotation angle on the operating table, while the Side camera is used to monitor object slipping during the grasping process and return the transformation matrix from the eelink frame to the object frame.

$\textbf{Top-down Camera}$: In order to generate a grasping pose, we obtain the RGBD information returned by the Top-down camera. For excluding the interference of background or irrelevant objects, we perform semantic segmentation using RGB images. Then we find out the smallest external rectangle representation that can surround the segmented image which can be further reduced to two vectors that are perpendicular to each other and traverse the center coordinates. Since we only consider the rod-shape object in this paper, such as a hammer or a wooden stick, it can be assumed that the grasping pose should mainly lie in the object ' length direction. Therefore,we use the vector with the larger modulus as the direction vector of the object and the center of the rectangle as its geometric center to represent the grasping pose in camera frame.Finally we can obtain the geometry-based grasping pose $(x,y,z,orientation)$ in world frame after using the transformation matrix from camera frame to world frame.

\begin{figure}[htbp]
\centering
\includegraphics[width=1\textwidth]{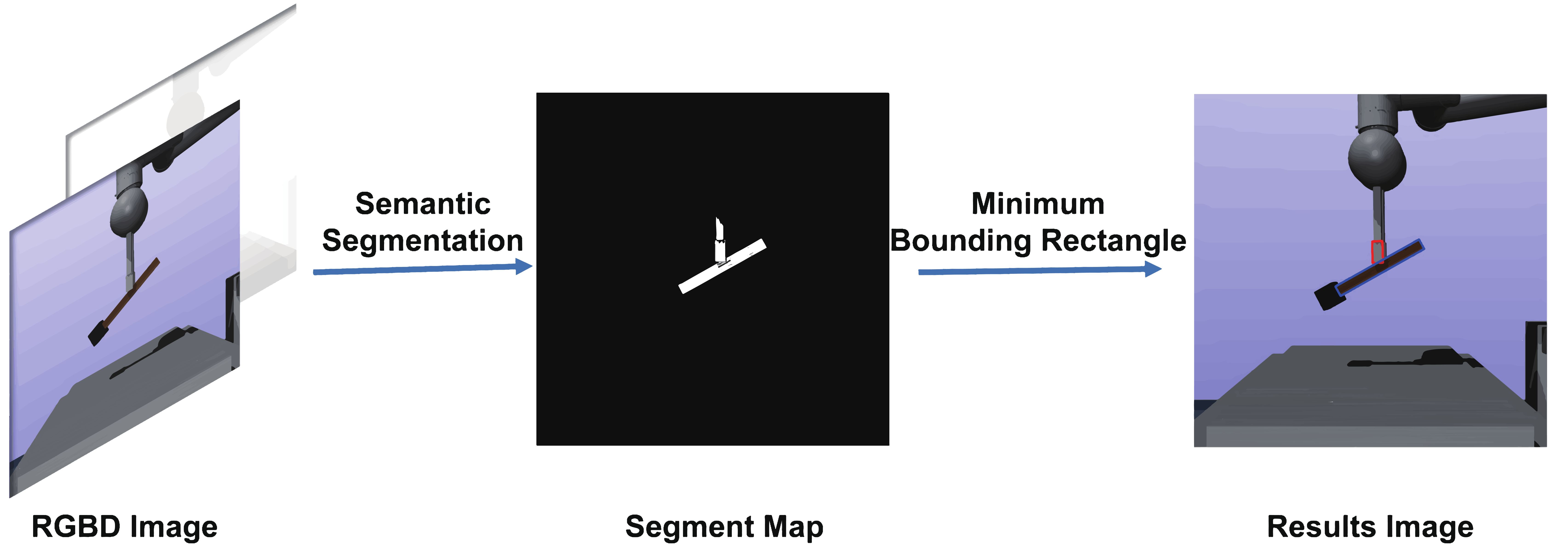}
\caption{Object posture estimation pipeline}
\label{fig:object posture estimation pipeline}
\end{figure}

$\textbf{Side Camera}$: Due to the uneven mass distribution of the object,  the moment generated by object ' gravity  may produce and then aggravate the object slipping, resulting in the object eventually sliding down to the table and the grasping failure. In the process of gripper lifting, the top-down camera cannot accurately monitor how the object would slip due to the occlusion of the robot  body, so we will adopt the strategy of adding cameras on the side of the 
eelink to make the perceived information more adequate and perfect. Similar to the Top-down camera, we semantically segment the RGBD information obtained by the Side camera and find the smallest external rectangle that can enclose the segmented image to obtain the direction vector and geometric center of gravity of the object as Fig.~\ref{fig:object posture estimation pipeline} illustrates. Considering that the gripper is also of slender type, we simplify the contact between the gripper and object to two intersection points and determine the transformation matrix $M$ from eelink frame to object frame by their displacement and rotation relationship.

\subsection{Estimating the CoM location in eelink frame}
In this section, we calculate the $(x,y)$ position of the object's CoM in eelink frame using hydrostatic knowledge based on the joint sensors.

\begin{figure}[!htbp]
\includegraphics[width=1\textwidth]{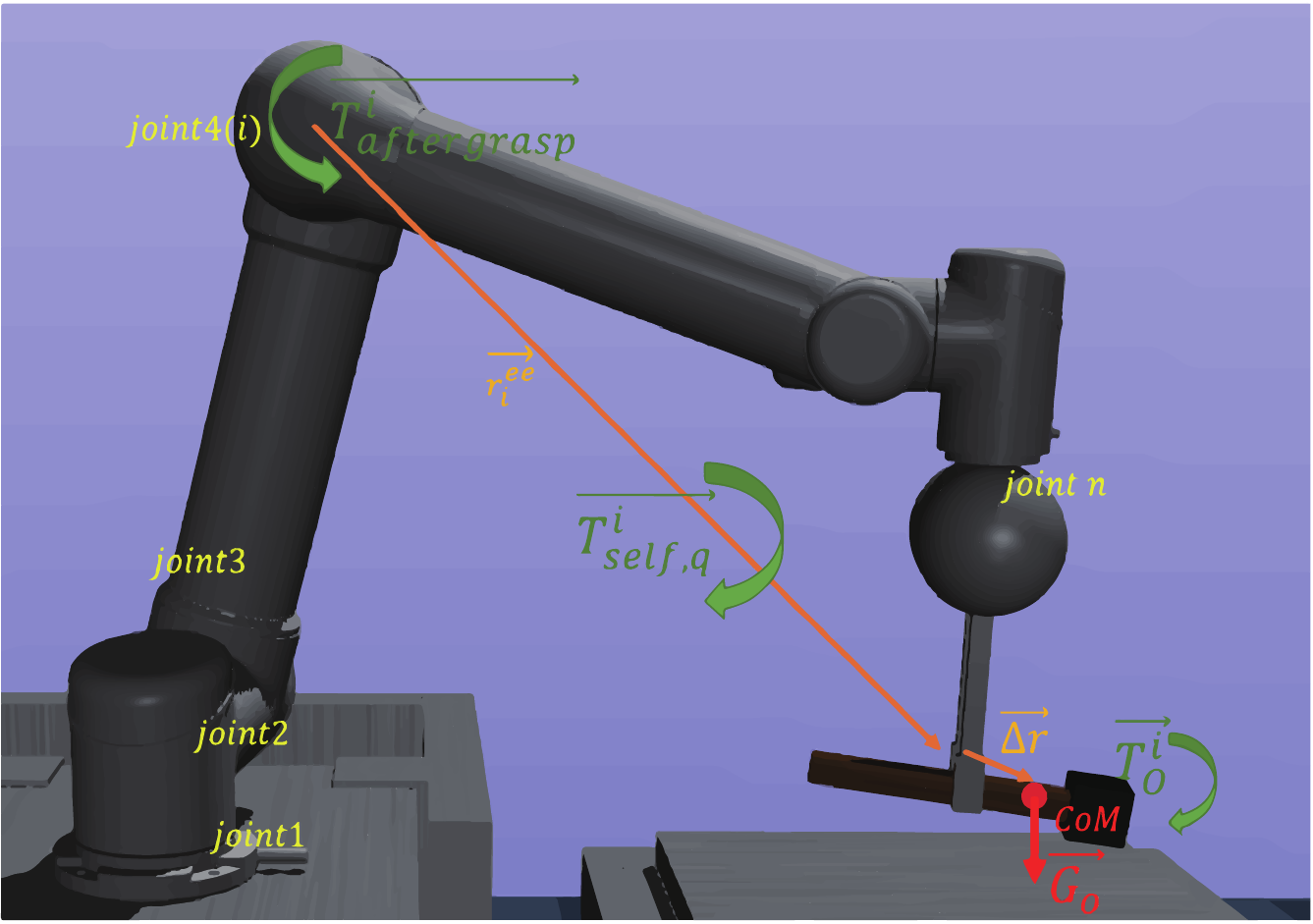}
\caption{We perform moment balance analysis for joint $i$.}
\label{fig:moment balance analysis}
\end{figure}

We consider a n-degree-of-freedom robot arm, numbering the axes at the base as $1$ and the remaining axes sequentially as $2,3,..n$.
We take the joints numbered $n,n-1,...i$ and their child links as a whole .When the arm maintains a status $q\in R^{n} $ which represents the arm state in joint space,
the whole part we focus is subject to three parts of torques as Fig.~\ref{fig:moment balance analysis} illustrates: $\overrightarrow{T_{self,q}^{i} }$ generated by its own gravity, $\overrightarrow{T_{O}^{i} }$ generated by the object gravity $\overrightarrow{G_{o} }$,   $\overrightarrow{T_{aftergrasp}^{i}}$ generated by the joint $i$.

We perform moment balance analysis for joint $i$.
\begin{equation}
    \overrightarrow{T_{O}^{i} }
+\overrightarrow{T_{aftergrasp}^{i} }+\overrightarrow{T_{self,q}^{i} }=0
\end{equation}

We denote the position vector from $i$ joint to eelink by $\overrightarrow{r_{i}^{ee}}$, the position vector from  eelink to the object ' CoM by $\overrightarrow{\Delta r}$ and rewrite equation (1) as:
\begin{equation}
    (\overrightarrow{r_{i}^{ee}}+  \overrightarrow{\Delta r})\times \overrightarrow{G_{o}} 
+\overrightarrow{T_{aftergrasp}^{i} }+\overrightarrow{T_{self,q}^{i} }=0
\end{equation}

However, considering the complexity of the actual robot arm structure, it is very difficult to directly calculate $\overrightarrow{T_{self,q}^{i} }$. In order to eliminate the influence of the arm's own gravity, we measure the torque value $\overrightarrow{T_{beforegrasp}^{i} } $ generated by the joint $i$ when the arm is not grasping the object in status $q$ and we have following equation:
\begin{equation}
     \overrightarrow{T_{beforergrasp}^{i} }+\overrightarrow{T_{self,q}^{i} }=0
\end{equation}

By combining equation (2) and (3), we have: 
\begin{equation}
    (\overrightarrow{r_{i}^{ee}}+  \overrightarrow{\Delta r})\times \overrightarrow{G_{o}} 
+\overrightarrow{T_{aftergrasp}^{i} }-\overrightarrow{T_{beforergrasp}^{i} }=0
\end{equation}

In reality, we can only get the active torque of joints along the direction of rotation, and we project both sides of the equation(4) in the direction of the axis of joint $i$.

\begin{equation}
    Proj_{\overrightarrow{Axis_{i} }} ((\overrightarrow{r_{i}^{ee}}+  \overrightarrow{\Delta r})\times \overrightarrow{G_{o}})+\overrightarrow{T_{aftergrasp}^{i,axis}}
-\overrightarrow{T_{beforegrasp}^{i,axis}}=0
\end{equation}

Since $\overrightarrow{Axis_{i} }$ , $\overrightarrow{r_{i}^{ee}}$ , $\overrightarrow{T_{beforergasp}^{i,axis} } $ , $\overrightarrow{T_{aftergrasp}^{i,axis} } $ can be obtained through joint position and torque sensors, we can simplify the task of solving position of object's CoM in eelink frame to a optimization problem.  

\begin{equation}
    \begin{cases}
s.t \quad i \in \left \{ 1,2,3...n \right \} 
\\
\underset{arg \vec{\Delta r} ,{G_{o}}}{min} \sum_{i=1}^{n}(Proj_{\overrightarrow{Axis_{i} }} ((\overrightarrow{r_{i}^{ee}}+  \overrightarrow{\Delta r})\times \overrightarrow{G_{o}})+\overrightarrow{T_{aftergrasp}^{i,axis}}
-\overrightarrow{T_{beforegrasp}^{i,axis}})^{2}
\end{cases}
\end{equation}

As our objective function is a kind of low order polynomial, we apply the gradient descent method in the algorithm to obtain $\overrightarrow{\Delta r}$ and $G_{o}$.
Considering the cross product process in (6) which ignores $\Delta r_{z} $ in $\overrightarrow{\Delta r}(\Delta r_{x},\Delta r_{y},\Delta r_{z} )$, we finally obtain {$\Delta r_{x},\Delta r_{y},G_{o} $} .$\Delta r_{x},\Delta r_{y}$ represents object's CoM $(x,y)$ position in eelink frame and $G_{o}$ represents the object's gravity.
\subsection{CoM estimation in object frame and regrasp planner}
\begin{figure}[!htbp]
\includegraphics[width=1\textwidth]{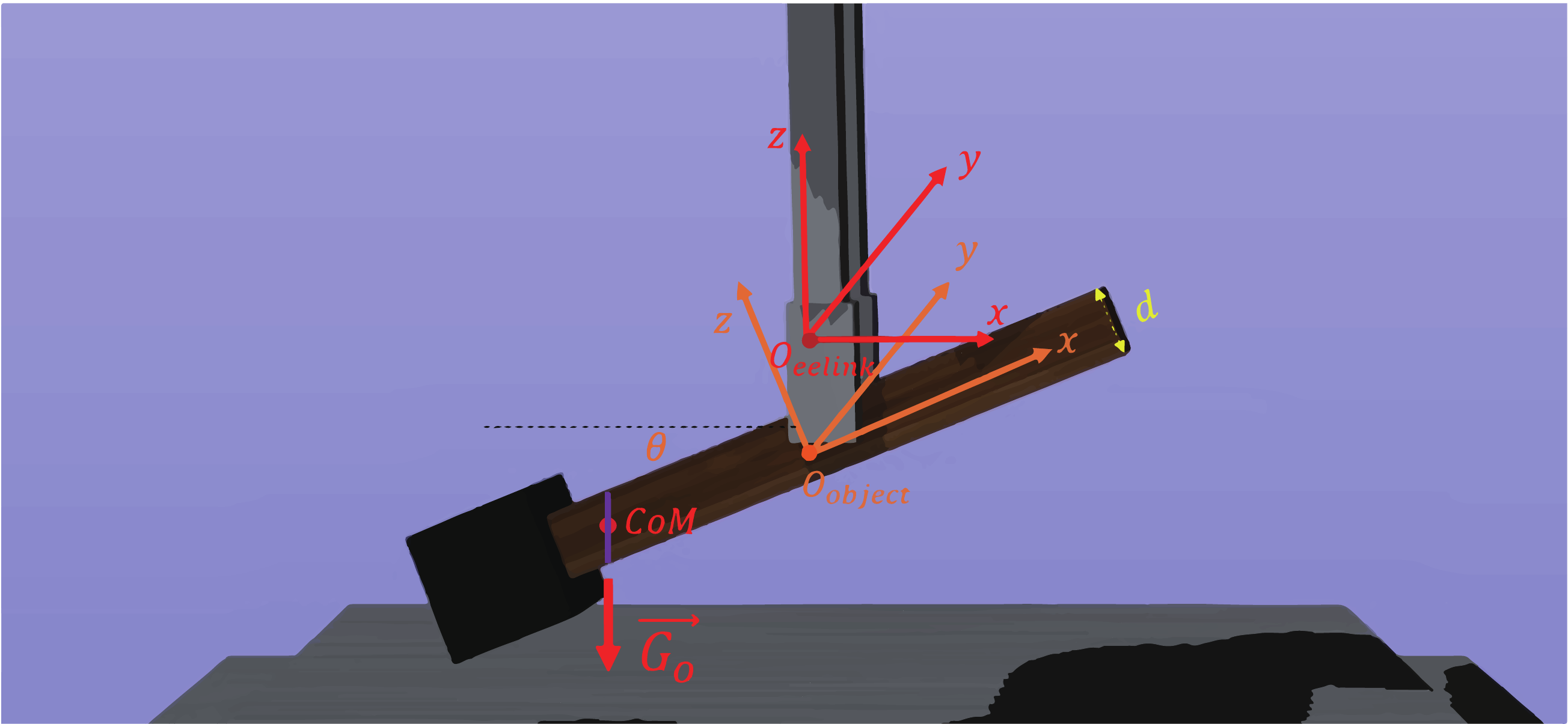}
\caption{The purple line segment represents where the CoM is lying}
\label{fig:The purple line segment represents where the CoM is lying}
\end{figure}
We have obtained the coordinate system transfer matrix $M$ from eelink frame to object frame in section 4.1, and the position of the object's CoM  $\overrightarrow{\Delta r}$ in eelink frame in section 4.2, so that we can calculate where the CoM locates in object frame by the following equation :
\begin{equation}
\begin{pmatrix}
x_{CoM}^{O} 
 \\
y_{CoM}^{O}
 \\
z_{CoM}^{O}
 \\
1
\end{pmatrix}=M\begin{pmatrix}
\overrightarrow{\Delta r}^{T}  
\\
1
\end{pmatrix}   =\begin{pmatrix}
 r_{11} &r_{12}  & r_{13}&p_{x}  \\
  r_{21}&  r_{21}& r_{21} & p_{y}\\
  r_{31}&  r_{31}&  r_{31}& p_{z}\\
  0& 0 & 0 &1
\end{pmatrix}\begin{pmatrix}
\Delta r_{x} 
\\\Delta r_{y} 
 \\
\Delta r_{z}^{*}
 \\
1
\end{pmatrix}
\end{equation}

Since $\Delta r_{z}^{*}$ is indeterminate, the location of the CoM solved is lying on a line segment inside the object, which is depicted as a purple line in Fig.~\ref{fig:The purple line segment represents where the CoM is lying}. The smaller both the cross-sectional radius $d$ and the tilt angle $\theta $ when it is lifted, the more accurate the location of the CoM will be. Considering the object is thin rod-shaped, we let $z_{CoM}^{O}=0$, and so that obtain $ x_{CoM}^{O}$: 
 \begin{equation}
x_{CoM}^{O}=\frac{r_{11}r_{33}-r_{13}r_{31}}{r_{33}} \Delta r_{x}+
\frac{r_{12}r_{33}-r_{13}r_{32}}{r_{33}} \Delta r_{y}+p_{x}-\frac{r_{13}}{r_{33}} p_{z} 
\end{equation}

Once we observe a slide in the first grasp trial, we can generate a new grasp pose based on $x_{CoM}^{O}$ so that the second grasp is very close to the CoM. 
\section{EXPERIMENT}
\subsection{Experimental setup}
\begin{figure}[!htbp]
\includegraphics[width=1\textwidth]{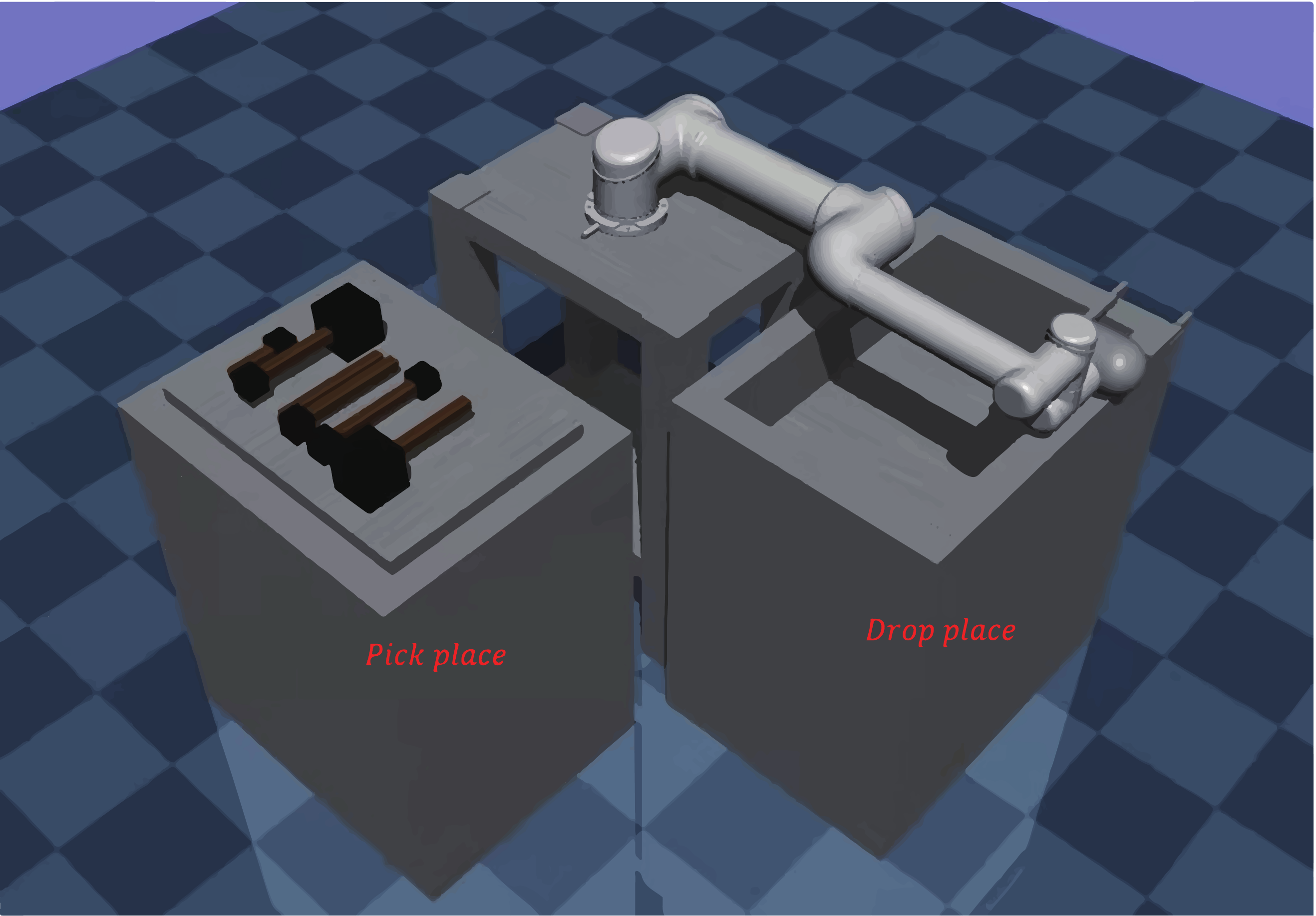}
\caption{The overview of experiment equipment}
\label{fig:the overview of experiment equipment}
\end{figure}

In this work, grasping experiments are conducted using a 6-DOFs UR5 robot arm in "Mujoco", equipped with a custom parallel-jaw gripper. A table as $pick\ box$ is placed in front of the arm while a $drop\ box$ is left of the arm. As shown in Fig.~\ref{fig:the overview of experiment equipment}, we prepare 6 different kinds of objects on the $pick\ box$. The experiments run on a laptop installed with Ubuntu 18.04 with a 2.21 GHz Intel Core i7-8750H 12-Core CPU and an NVIDIA GeForce GT 1060 graphic card. 

The UR5 parallel gripper can be controlled to close with specific force. Two "Mujoco" built-in RGBD cameras are mounted at the height of 200 cm above the ground and at the distance of 20 cm left of $pick\ box$. We capture the images from two cameras during every simulation step.

\subsection{Accuracy of estimated CoM location test}
In "Mujoco", we can modify the object parameters in MJCF to set the CoM position $pos_{real}$ and the object length $l$.During each grasp, we collect  object tilt angle $\theta$ and joint position and torque sensors data to estimate CoM position $pos_{esti}$ using our proposed method in section4 .$1-5\left | \frac{pos_{esti}-pos_{real}}{l} \right |  $ is used to measure the accuracy of the estimated CoM position. 

We conducted 50 grasping experiments using random grasp pose based on the rectangle representation for each of the 6 different objects.

From Fig.~\ref{fig:CoM Location Accuracy}, we can see that our CoM estimation algorithm has good accuracy when the object tilt angle is relatively small.

What’s more ,there are two important messages delivered from Fig.~\ref{fig:CoM Location Accuracy}:

1. The estimation accuracy decreases rapidly after a threshold value of $\theta$.

2. The thicker the object, the lower the estimation accuracy.

The reason for these two phenomena is exactly what we have analyzed in Section 4.3: since the calculated center-of-mass location is on a line segment inside the object, the uncertainty of CoM's estimated location will grow as the object tilt angle $\theta$ and the cross-section radius $d$ increase.In addition,the complex slipping dynamics may also take part in reducing the estimation accuracy.

\subsection{Regrasp stability test}

We test the effectiveness of our proposed method on improving the grasp stability in this experiment. We design a grasping task: the robotic arm grasps an object on $pick\ box$ and places it on $drop\ box$. If the grasping is unstable, the object is likely to fall during the movement of the robotic arm. We respectively conduct experiments with RGBD only based method and our proposed method.

\begin{figure}[htbp]
    \centering
    \subfigure[CoM Location Accuracy]{
    	\label{fig:CoM Location Accuracy}
        \begin{minipage}[t]{0.5\linewidth}
            \centering
            \begin{tikzpicture}[scale=0.7]
                \centering
                \begin{axis}[
                    xlabel=$\theta$,
                    ylabel=Accuracy,
                    axis lines*=left,
                    ymajorgrids = true,
                    xmajorgrids = true,
                    tick align=outside,
                    legend style={at={(0.5,-0.2)},anchor=north,legend columns=-1},
                ]
                    \addplot[draw=ProcessBlue,mark=star] 
                    coordinates {
                        (0.12,98.68)
                        (1.35,98.98)
                        (2.65,98.69)
                        (2.98,98.62)
                        (3.25,97.69)
                        (4.55,97.54)
                        (5.68,97.50)
                        (7.36,97.03)
                        (8.65,96.89)
                        (9.15,96.79)
                        (10.35,95.21)
                        (12.36,93.56)
                        (13.56,87.65)
                        (16.59,83.36)
                        (20.65,78.68)
                        (25.46,73.56)
                        (30.79,70.36)
                    };
                    \addlegendentry{obj1}
                    
                    \addplot[draw=BrickRed,mark=square] 
                    coordinates {
                        (0.12,98.32)
                        (1.35,97.98)
                        (2.65,97.69)
                        (2.98,97.62)
                        (3.25,96.52)
                        (4.55,96.50)
                        (5.68,96.49)
                        (7.36,96.01)
                        (8.65,95.92)
                        (9.15,94.82)
                        (10.35,91.21)
                        (12.36,85.64)
                        (13.56,83.96)
                        (16.59,82.52)
                        (20.65,75.68)
                        (25.46,71.56)
                        (30.79,69.35)
                    };
                    \addlegendentry{obj2}
                    
                    \addplot[draw=YellowGreen,mark=triangle] 
                    coordinates {
                        (0.15,97.32)
                        (1.38,96.98)
                        (2.69,96.69)
                        (2.94,96.62)
                        (3.21,95.52)
                        (4.50,95.50)
                        (5.69,95.49)
                        (7.31,95.01)
                        (8.69,92.92)
                        (9.11,92.82)
                        (10.31,89.02)
                        (12.30,85.64)
                        (13.59,83.96)
                        (16.54,81.02)
                        (20.61,74.01)
                        (25.40,70.42)
                        (30.72,68.32)
                    };
                    \addlegendentry{obj3}

                    \addplot[draw=Melon,mark=x] 
                    coordinates {
                        (0.15,96.32)
                        (1.38,96.21)
                        (2.69,95.02)
                        (2.94,94.62)
                        (3.21,94.21)
                        (4.50,94.03)
                        (5.69,93.98)
                        (7.31,93.65)
                        (8.69,93.60)
                        (9.11,93.32)
                        (10.31,86.98)
                        (12.30,83.12)
                        (13.59,81.49)
                        (16.54,79.00)
                        (20.61,71.89)
                        (25.40,68.95)
                        (30.72,66.21)
                    };
                    \addlegendentry{obj4}

                    \addplot[draw=OliveGreen,mark=+] 
                    coordinates {
                        (0.15,94.20)
                        (1.38,94.14)
                        (2.69,94.06)
                        (2.94,92.52)
                        (3.21,92.02)
                        (4.31,92.00)
                        (5.18,88.65)
                        (7.16,88.14)
                        (8.14,88.60)
                        (9.01,87.32)
                        (10.31,85.45)
                        (12.30,82.12)
                        (13.15,80.49)
                        (16.54,76.00)
                        (20.61,72.89)
                        (25.40,68.86)
                        (30.72,62.19)
                    };
                    \addlegendentry{obj5}

                    \addplot[draw=Orchid,mark=diamond] 
                    coordinates {
                        (0.15,90.20)
                        (1.38,90.14)
                        (2.69,90.06)
                        (2.94,89.52)
                        (3.21,88.02)
                        (4.31,86.00)
                        (5.18,85.65)
                        (7.19,85.14)
                        (8.16,85.60)
                        (9.33,85.32)
                        (10.98,82.45)
                        (12.14,78.12)
                        (13.68,75.49)
                        (16.13,74.00)
                        (20.48,65.89)
                        (25.68,63.86)
                        (30.56,59.05)
                    };
                    \addlegendentry{obj6}
                \end{axis}
            \end{tikzpicture}
        \end{minipage}%
    }%
    \subfigure[Regrasp Stability Test]{
        \label{fig:Regrasp Stability test}
        \begin{minipage}[t]{0.5\linewidth}
            \centering
            \begin{tikzpicture}[scale=0.7]
                \centering
            	\begin{axis}[
            	    symbolic x coords={obj1, obj2, obj3, obj4, obj5, obj6, 0},
            	    xtick={obj1, obj2, obj3, obj4, obj5, obj6},
            	    x tick label style={rotate=45, anchor=east, align=center},
            	    ylabel=Grasp Success Rate,
            	    axis lines*=left,
            	    ymajorgrids = true,
            	    ybar interval = 0.8,
            	    ymin = 0,
            	    legend style={at={(0.5,-0.2)},anchor=north,legend columns=-1},
            	]
                	\addplot[draw=ProcessBlue, fill=ProcessBlue] 
                	coordinates{
                		(obj1,93.2)
                        (obj2,91.3)
                        (obj3,90.5)
                        (obj4,89.1)
                        (obj5,87.6)
                        (obj6,86.9)
                        (0,0)
                	};
                	\addplot[draw=BrickRed, fill=BrickRed] 
                	coordinates{
                		(obj1,70.8)
                        (obj2,65.6)
                        (obj3,61.5)
                        (obj4,60.6)
                        (obj5,58.5)
                        (obj6,58.0)
                        (0,0)
                	};
                	\addlegendentry{Ours regrasp planner}
                    \addlegendentry{Without Ours Methods}
                \end{axis}
            \end{tikzpicture}
        \end{minipage}%
    }%

    \caption{Experiment results}
\end{figure}
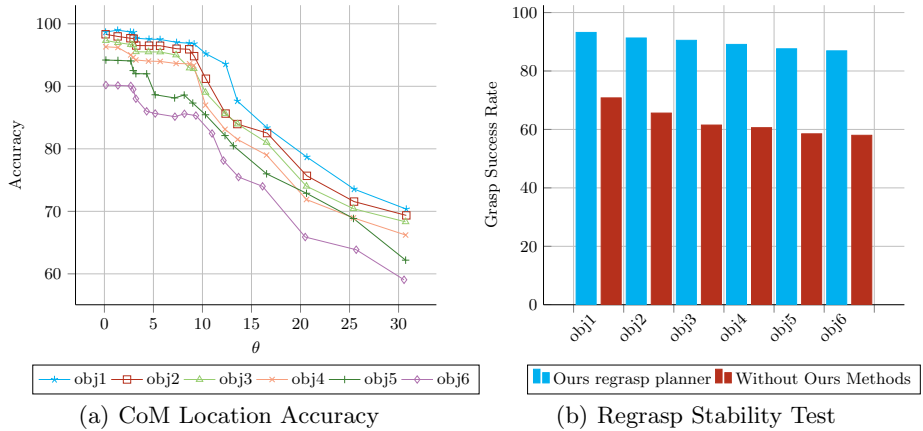

From Fig.~\ref{fig:Regrasp Stability test}, we can see that our method effectively improves the grasping stability for rod-shaped objects.

\section{CONCLUSION}
In this paper, we presented an algorithm that can obtain the center-of-mass position of the rod object using the widely fixed joint torque sensors on the robot arm and the RGBD camera on the side of the robot jaw. Our method does not require additional wrist moment sensors or additional acquisition of large amounts of experimental data.

In future work, we first plan to enhance the grasped objects from linear to non-linear so as to better match the requirements of natural objects. We also plan to fuse data from multiple sources, such as tactile sensors with joint torques, to improve the accuracy of our algorithm further.

\section*{Acknowledgement}
\addcontentsline{toc}{section}{Acknowledgement}
This research was funded by National Natural Science Foundation of China (No.61603020) and the Fundamental Research Funds for the Central Universities (No.YWF-21-BJ-J-923). 
%
%

\bibliographystyle{splncs03_unsrt}

\bibliography{author}

\begin{thebibliography}{1}
\providecommand{\url}[1]{\texttt{#1}}
\providecommand{\urlprefix}{URL }

\bibitem{GPD}
ten Pas, A., Gualtieri, M., Saenko, K., Jr., R.P.: Grasp pose detection in
  point clouds. CoRR  abs/1706.09911 (2017),
  \url{http://arxiv.org/abs/1706.09911}

\bibitem{Efficient-grasping}
Jiang, Y., Moseson, S., Saxena, A.: Efficient grasping from rgbd images:
  Learning using a new rectangle representation. In: 2011 IEEE International
  Conference on Robotics and Automation. pp. 3304--3311 (2011)

\bibitem{DL-Detecting-Robotic-Grasps}
Lenz, I., Lee, H., Saxena, A.: Deep learning for detecting robotic grasps. CVPR
   (2014), \url{http://arxiv.org/abs/1301.3592}

\bibitem{Inertial-Properties}
Lopez-Damian, E., Sidobre, D., Alami, R.: A grasp planner based on inertial
  properties. In: Proceedings of the 2005 IEEE International Conference on
  Robotics and Automation. pp. 754--759 (2005)

\bibitem{4048449}
Atkeson, C.G., An, C.H., Hollerbach, J.M.: Rigid body load identification for
  manipulators. In: 1985 24th IEEE Conference on Decision and Control. pp.
  996--1002 (1985)

\bibitem{slip-detection-strategies}
Reinecke, J., Dietrich, A., Schmidt, F., Chalon, M.: Experimental comparison of
  slip detection strategies by tactile sensing with the biotac on the dlr hand
  arm system. In: 2014 IEEE International Conference on Robotics and Automation
  (ICRA). pp. 2742--2748 (2014)

\bibitem{Slip-Detection}
Wyk, K.V., Falco, J.: Slip detection: Analysis and calibration of univariate
  tactile signals. CoRR  abs/1806.10451 (2018),
  \url{http://arxiv.org/abs/1806.10451}

\bibitem{center-of-mass}
Kanoulas, D., Lee, J., Caldwell, D.G., Tsagarakis, N.G.: Center-of-mass-based
  grasp pose adaptation using 3d range and force/torque sensing. CoRR
  abs/1802.06392 (2018), \url{http://arxiv.org/abs/1802.06392}

\bibitem{Center-of-Mass-based}
Feng, Q., Chen, Z., Deng, J., Gao, C., Zhang, J., Knoll, A.C.:
  Center-of-mass-based robust grasp planning for unknown objects using
  tactile-visual sensors. CoRR  abs/2006.00906 (2020),
  \url{https://arxiv.org/abs/2006.00906}

\end{thebibliography}

\end{document}